\documentclass[letterpaper]{article} 
\usepackage{aaai2026}  
\usepackage{times}  
\usepackage{helvet}  
\usepackage{courier}  
\usepackage[hyphens]{url}  
\usepackage{graphicx} 
\usepackage{natbib}  
\usepackage{caption} 
\usepackage{algorithm}
\usepackage{algorithmic}

\usepackage{newfloat}
\usepackage{listings}
\DeclareCaptionStyle{ruled}{labelfont=normalfont,labelsep=colon,strut=off} 
\floatstyle{ruled}
\newfloat{listing}{tb}{lst}{}
\floatname{listing}{Listing}
\usepackage{amsmath}
\usepackage{adjustbox}
\usepackage{subcaption}
\usepackage{array}
\usepackage{booktabs}       
\usepackage{amsfonts}       
\usepackage{nicefrac}       
\usepackage{microtype}      
\usepackage{xcolor}         

\title{You Only Need One Stage: Novel-View Synthesis From A Single Blind Face Image}

\author {
    Taoyue Wang\textsuperscript{\rm 1},
    Xiang Zhang\textsuperscript{\rm 1},
    Xiaotian Li\textsuperscript{\rm 1},
    Huiyuan Yang\textsuperscript{\rm 2},
    Lijun Yin\textsuperscript{\rm 1}
}
\affiliations {
    \textsuperscript{\rm 1}School of Computing, State University of New York at Binghamton\\
    \textsuperscript{\rm 2}Department of Computer Science, Missouri University of Science and Technology\\
    twang61@binghamton.edu, zxiang4@binghamton.edu, xli210@binghamton.edu, hyang@mst.edu, lijun@cs.binghamton.edu
}

\makeatletter
\renewcommand{\@seccntformat}[1]{
  \csname the#1\endcsname \hspace{0.3em}
}
\makeatother

\begin{document}

\maketitle

\begin{abstract}
We propose a novel one-stage method, \textbf{NVB-Face}, for generating consistent \textbf{N}ovel-\textbf{V}iew images directly from a single \textbf{B}lind \textbf{Face} image. Existing approaches to novel-view synthesis for objects or faces typically require a high-resolution RGB image as input. When dealing with degraded images, the conventional pipeline follows a two-stage process: first restoring the image to high resolution, then synthesizing novel views from the restored result. However, this approach is highly dependent on the quality of the restored image, often leading to inaccuracies and inconsistencies in the final output. To address this limitation, we extract single-view features directly from the blind face image and introduce a feature manipulator that transforms these features into 3D-aware, multi-view latent representations. Leveraging the powerful generative capacity of a diffusion model, our framework synthesizes high-quality, consistent novel-view face images. Experimental results show that our method significantly outperforms traditional two-stage approaches in both consistency and fidelity.
\end{abstract}


\section{Introduction}
\label{sec:intro}

Reconstructing 3D structures or generating novel-view images from a single face image has long been a fundamental research topic, with applications spanning digital human modeling and 3D animation. Humans can effortlessly imagine different viewpoints of a given face due to their extensive prior knowledge of facial structures and multi-view consistency. However, in computer vision, faithfully synthesizing novel views while preserving the exact identity, expression, background, and other attributes remains an extremely challenging and ill-posed problem.

\begin{figure}[t]
\centering
\includegraphics[width=\linewidth]{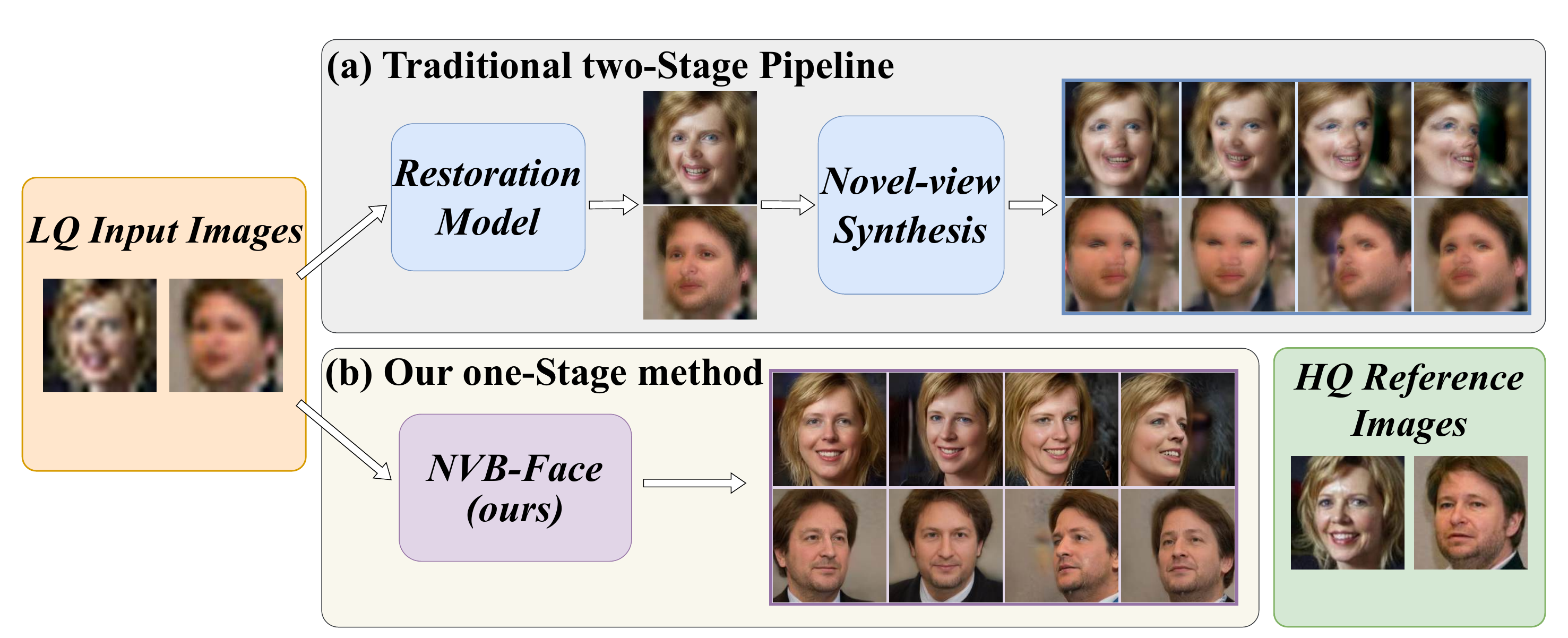}
\caption{We compare our method with typical two-stage pipelines, such as CodeFormer~\cite{zhou2022towards} + PanoHead-PTI~\cite{an2023panohead}, which first restore the degraded image and then synthesize novel views. It is evident that when the restoration stage fails to recover accurate details, these errors are further amplified during the novel view synthesis, leading to results that deviate significantly from the original identity and appearance. In contrast, our method generates novel views in a single stage directly from the low-quality input. This end-to-end design suppresses error accumulation, resulting in more reliable and faithful novel-view images. 
} 
\label{fig:first}
\end{figure}

Currently, mainstream approaches for novel-view face generation fall into three major categories: regression-based models such as 3D Morphable Models (3DMM)~\cite{paysan20093d, wu2019mvf, deng2019accurate, dib2021towards, zhuang2022mofanerf}, Neural Radiance Fields (NeRF)-based methods~\cite{chan2021pi, gu2021stylenerf, zhang2022fdnerf, chan2022efficient, yin20233d, yuan2023make, an2023panohead, trevithick2023real, bhattarai2024triplanenet}, and generative methods leveraging Stable Diffusion and ControlNet~\cite{rombach2022high, zhang2023adding, gu2024diffportrait3d, gu2025diffportrait360}. Despite their differences, these methods share a critical limitation—\textbf{they all require a high-quality, single-view face image as input}. Unfortunately, obtaining high-quality face images is not always feasible, as most images inevitably suffer from some degree of degradation. When dealing with low-resolution, blurry, noisy, or compression-degraded face images, the conventional approach involves a two-stage pipeline: first restoring the input image and then performing novel-view synthesis. However, this two-stage inference process introduces a fundamental issue: the performance of novel-view generation is entirely dependent on the quality of the restored image. Since the two stages operate independently without explicit connections, any errors in the upstream image restoration step get amplified in the downstream novel-view synthesis, significantly degrading the final output quality (as shown in Figure~\ref{fig:first}). Moreover, such a pipeline is inherently inefficient. Novel-view synthesis can only be performed after obtaining and evaluating the restoration results, often requiring additional selection or filtering to ensure quality. This dependency introduces a major bottleneck for large-scale deployment, making it less practical for real-world applications.

To address this issue, we propose an end-to-end framework based on Stable Diffusion~\cite{rombach2022high} that enables the direct generation of high-quality novel-view images from a single blind face image in a single inference stage. Our overall pipeline begins with extracting features from a low-quality input face image. These features are then transformed into novel-view feature representations corresponding to the target camera viewpoint. Finally, we leverage the strong generative capability of Stable Diffusion to reconstruct high-resolution novel-view images from the transformed low-quality features. Specifically, to enable viewpoint transformation in latent space, we introduce a Transformer-based 3D Feature Construction Module, which builds a latent 3D face representation grid from the single-view input. By leveraging camera parameters, we are able to directly project this 3D representation into novel-view features within the latent space.
It is important to highlight that, unlike previous diffusion-based novel-view synthesis methods~\cite{papantoniou2024arc2face, gu2024diffportrait3d, gu2025diffportrait360} that rely on ControlNet~\cite{zhang2023adding}, where camera parameters are first used to generate image templates as conditional inputs, our method directly performs viewpoint-conditioned mapping on the generated 3D feature grid using the camera parameters. This not only provides a more intuitive and flexible modeling strategy, but also significantly improves multi-view consistency due to the use of an explicit 3D feature representation. Furthermore, our pipeline employs a single-stage, end-to-end inference process, which avoids the error accumulation commonly found in multi-stage approaches, resulting in higher computational efficiency and stronger generalization.

Our contributions can be summarized as follows.
\begin{enumerate}
\item We propose a novel tuning-free framework that directly generates high-quality novel-view face images from a single blind face input at specified viewpoints. 
To the best of our knowledge, our approach is the first work to explore this end-to-end manner.
\item We introduce a novel 3D latent space representation of facial features, which enables consistent and accurate multi-view feature projection. This structured representation helps maintain cross-view consistency during novel-view synthesis.
\item Extensive qualitative and quantitative experiments demonstrate that our one-stage novel-view synthesis method achieves state-of-the-art performance, validating that our framework produces superior perceptual quality compared to traditional two-stage pipelines.
\end{enumerate}

\section{Related Work}
Our proposed method integrates two challenging tasks: \textbf{blind face restoration} and \textbf{novel view synthesis}. In this section, we provide a brief overview of related works in both areas and discuss why directly combining these two tasks into a single inference stage for generating high-quality novel views from a degraded image is inherently difficult.
\paragraph{Blind Image Restoration.} 
Restoring a degraded image into a high-resolution, detail-rich version is an inherently ill-posed problem. Over the past decade, researchers have leveraged the powerful generative capabilities of deep models, particularly GANs~\cite{yang2021gan, dib2021towards, wang2021towards, wang2022restoreformer, zhou2022towards, wang2023restoreformer++} and diffusion models~\cite{wang2023dr2, yue2023resshift, wang2024exploiting, lin2024diffbir, yue2024difface, sun2024improving, yue2023resshift, wu2025one, wang2024osdface}, to reconstruct missing details in degraded images.
GAN-based methods typically employ a U-Net-style generator to map degraded images to high-quality outputs, while a discriminator is used to enforce realism. However, GANs suffer from two major limitations. First, they are notoriously difficult to train, as maintaining the balance between the generator and discriminator remains a long-standing challenge. Second, their performance heavily depends on large-scale training data, often requiring millions of samples to generalize well. To address these issues, recent works have explored diffusion-based approaches for image restoration. Early diffusion-based methods~\cite{wang2023dr2, lin2024diffbir, yue2024difface, sun2024improving} for blind image restoration typically condition on the degraded image, using modules such as ControlNet~\cite{zhang2023adding} or a similar image encoder~\cite{wang2024exploiting} to extract features and inject them into a frozen pretrained Stable Diffusion model. The strong generative capability of Stable Diffusion then reconstructs the high-quality image. 
Given the compelling advantages of diffusion models, we also adopt a diffusion-based backbone. However, instead of following existing diffusion-based image restoration paradigms which require to first restore a single-view high-resolution image before performing a separate multi-view synthesis step, we propose a novel pipeline specifically designed to integrate seamlessly with the task of novel-view face synthesis. 

\begin{figure*}
\centering
\includegraphics[width=0.96\textwidth]{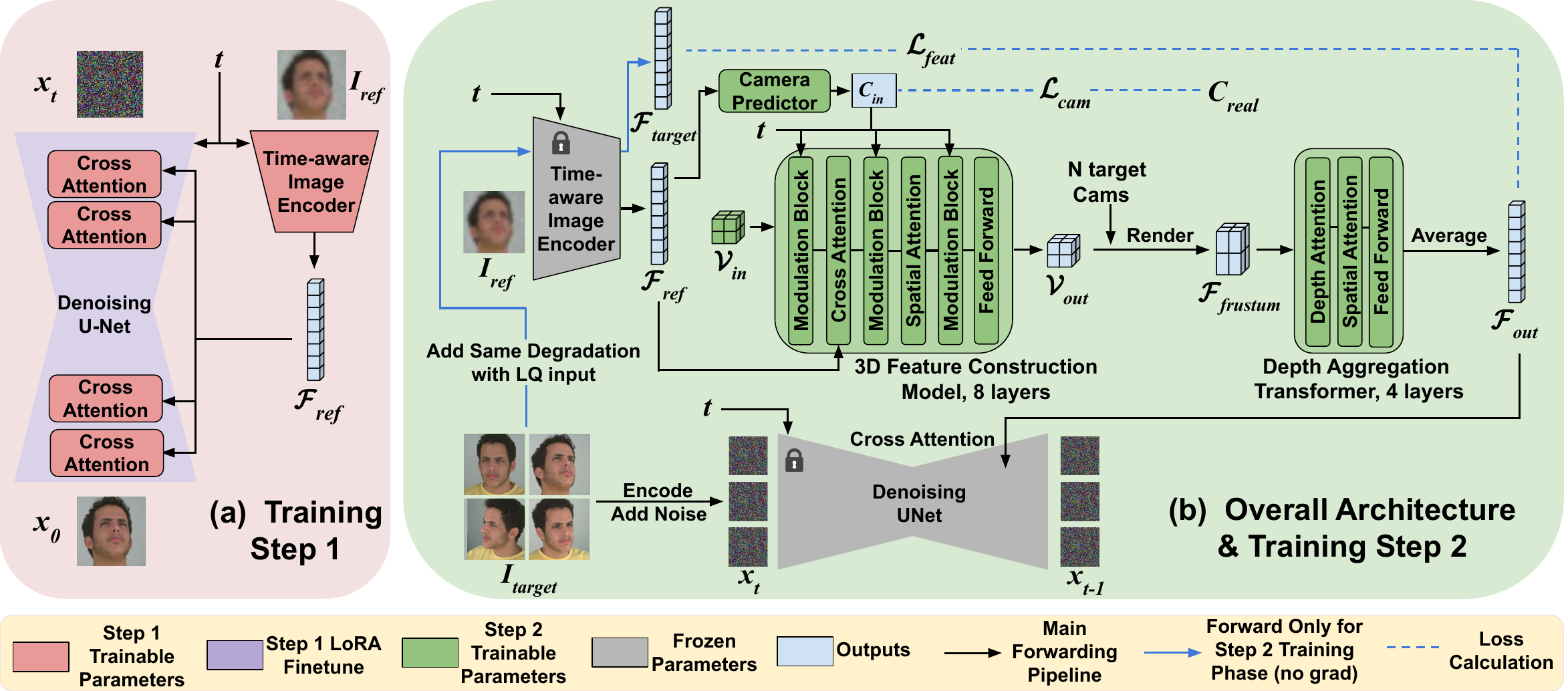}
\caption{An overview of the proposed \textbf{NVB-Face} architecture. \textbf{(a)} Our first training step focuses solely image restoration. \textbf{(b)} In the second training step, we update only the parameters of the newly introduced modules (highlighted in dark green), keeping the rest of the network frozen. After training, this two-step process forms our complete inference pipeline.}
\label{fig:arch}
\end{figure*}

\paragraph{Novel-View Face Synthesis.} Traditional regression-based models, such as 3D Morphable Models (3DMM)~\cite{paysan20093d, wu2019mvf, deng2019accurate, dib2021towards, zhuang2022mofanerf}, estimate parametric geometric priors of human faces and synthesize novel views by rendering 3D facial reconstructions. However, due to the limited expressiveness of the parameter space, 3DMM-based methods struggle to capture rich identity variations, expressions, and other facial attributes. As a result, researchers proposed NeRF-based methods~\cite{chan2021pi, gu2021stylenerf, zhang2022fdnerf, trevithick2023real} or their more efficient evolution, triplane-based representations~\cite{chan2022efficient, bhattarai2024triplanenet, an2023panohead, yin20233d, yuan2023make}. Among these, EG3D~\cite{chan2022efficient} was the first method to leverage 3D-aware GANs to generate a latent triplane representation and render high-quality, multi-view consistent face images. Following this work, a series of studies~\cite{yin20233d, yuan2023make, trevithick2023real, bhattarai2024triplanenet} explored GAN inversion techniques to fit EG3D's framework. However, all of these approaches rely on first extracting the camera viewpoint information from the input image to ensure consistency with EG3D's triplane representation. If the input image is degraded, extracting accurate camera parameters becomes highly challenging, making it necessary to first restore the image before novel-view synthesis — a process that inevitably introduces error accumulation and dependency on restoration quality. More recently, ~\cite{gu2024diffportrait3d} introduced the first diffusion-based approach for novel-view face generation. However, our experiments reveal a major limitation of this approach that while it works well for high-quality input images, it fails significantly for degraded inputs, leading to severe information loss.

To overcome these challenges, we propose a single-stage framework that seamlessly integrates blind face restoration and novel-view synthesis into a unified process. Unlike previous methods, our approach is able to directly take any face image of arbitrary quality and generate high-resolution and consistent novel-view outputs at specified camera viewpoints.

\section{Method}

Given an input image $\mathcal{I}_{ref}$ with an arbitrary level of degradation, our goal is to reconstruct the original details of $\mathcal{I}_{ref}$ while generating a high-resolution image that preserves the same identity, expression, and attributes, but from the specified new camera viewpoint $\mathcal{C}^{i}$. To achieve this, we propose NVB-Face, a framework that is trained in two steps. First, an image encoder extracts latent features $\mathcal{F}_{ref}$ from $\mathcal{I}_{ref}$, which are then fed into a Stable Diffusion (SD) model (Section~\ref{sec:subsection3.1}). Leveraging its strong generative capabilities, the SD model restores fine details and reconstructs a high-quality version of the degraded image while maintaining identity consistency. Next, we introduce a 3D-aware viewpoint transformation model that takes the single-view feature $\mathcal{F}_{ref}$ and transforms it into new features $\mathcal{F}_{out}^{i}$ according to the ${i}\text{th}$ camera viewpoints $\mathcal{C}^{i}$. The transformed features are then passed into our pretrained SD model to synthesize high-resolution novel-view images (Section~\ref{sec:subsection3.2}). Finally, we design a tailored loss function that effectively bridges blind face restoration and novel view synthesis into a unified training objective. We further provide theoretical and empirical justifications for how this loss formulation ensures seamless optimization of both tasks within a single-stage inference pipeline (Section~\ref{sec:subsection3.3}).

\subsection{Image Restoration} \label{sec:subsection3.1}
As depicted in Figure~\ref{fig:arch}a, given an low-quality (LQ) image, the restoration component of our NVB-Face first employs an image encoder $Enc$ to extract the feature $\mathcal{F}_{ref}$ of reference image $\mathcal{I}_{ref}$, which are then fed into the SD model through cross-attention to generate a high-quality image. 
It is important to note that we did not apply average pooling to the output features of the last layer. Instead, we retained the original feature \( \mathcal{F}_{ref} \in \mathbb{R}^{H \times W \times C} \) dimensions before feeding them into the Cross-Attention module. Since our task is image restoration, keeping the full latent spatial resolution of the features ensures that more fine-grained details from the input image are maintained, in order to faithfully reconstruct high-resolution outputs that accurately preserve the original contents.

Following the approach of~\cite{liu2023syncdreamer, wang2024exploiting}, we integrate temporal information into the image encoder to enhance generation quality and training stability. We incorporate the same time step embedding used in SD model into each image encoder block, making the image encoder time-aware, to ensure that the extracted features are synchronized with each time step of the downstream SD model.
\begin{align}
\mathcal{F}_{ref} = Enc(\mathcal{I}_{ref}, \; t)
\end{align}

Since we replace the original CLIP~\cite{radford2021learning} text encoder in the SD model with our custom image encoder, the existing cross-attention module, which was designed for text conditioning, is no longer effective. To address this, we jointly fine-tune both image encoder and the cross-attention parameters. Additionally, to make the model better suited for our specific task while preserving the inherent generative capabilities of the SD model, we fine-tune all components of the SD model using LoRA~\cite{hu2022lora, luo2023lcm}, except for the cross-attention module.

The training objective remains consistent with the original SD model, where the model is trained to predict noise at each time step. 
\begin{equation} \label{eq:sdloss}
    \begin{aligned}
    \mathcal{L}_{SD} = \mathbb{E}_{x_0, \mathcal{F}_{ref}, t, \epsilon \sim \mathcal{N}(0, 1)} \left[ \left\| \epsilon - \epsilon_\theta(x_t, t, \mathcal{F}_{ref}) \right\|^2 \right],
    \end{aligned}
\end{equation}
where $x_0$ is the latent image encoded by VAE encoder, and is diffused $t$ time steps into a Gaussian-distributed $x_t \sim \mathcal{N}(0, 1)$. And the $\epsilon_\theta$ is a U-Net structured noise predictor.

In addition to using conventional high-quality face dataset, we also leverage multi-view face datasets during the first training step. This ensures that, regardless of the subject ID or viewpoint of the features generated in the second step, the SD model can consistently reconstruct high-resolution images with accurate identity and detail preservation. Details about the datasets used in our experiments will be discussed in Section~\ref{sec:subsection4.1}.

\subsection{Novel View Synthesis} \label{sec:subsection3.2}

\paragraph{3D Feature Construction Model.}
To generate novel-view images of $\mathcal{I}_{ref}$, previous approaches~\cite{gu2024diffportrait3d, papantoniou2024arc2face} have attempted to synthesize a template image from the target viewpoint and feed it into ControlNet~\cite{zhang2023adding} to condition the model with camera parameters. However, there are two major issues. First, in our case, the SD model has already been fine-tuned during the first training step specifically for the image restoration task, which may partially compromise its generative capability for other tasks such as novel view generation. Second, ControlNet cannot inherently guarantee multi-view consistency, necessitating the integration of a trainable View Consistency Module within the SD model~\cite{gu2024diffportrait3d}, thereby increasing the overall model complexity.

Inspired by~\cite{peebles2023scalable} and~\cite{hong2023lrm}, We propose a transformer-based \textbf{3D Feature Construction Model}, denoted as $Trans$, which transforms the single-view feature $\mathcal{F}_{ref}$ extracted from the image encoder into a 3D feature volume $\mathcal{V}_{out}$ that fuses multi-view information, in order to enforces multi-view consistency in 3D space.
\begin{equation} \label{eq:trans}
    \begin{gathered}
        \mathcal{V}_{out} = Trans(\mathcal{V}_{in}, \mathcal{F}_{ref}, \mathcal{C}_{in}, t), \\
        \mathcal{C}_{in} = CameraPredictor(\mathcal{F}_{ref})
    \end{gathered}
\end{equation}
where $\mathcal{C}_{in}$ represents the camera parameters of the input image, while $t$ denotes the time step embedding, which is synchronized with the image encoder and the SD model. This design is inspired by~\cite{hong2023lrm}, which advocates constructing a 3D representation by leveraging both image features and corresponding camera parameters. The primary objective is to disentangle facial context (e.g., identity, expression) from pose, thereby enabling a more stable and consistent 3D feature representation. However, during inference, it is often impractical to obtain accurate viewpoint information from degraded input images. To address this issue, we introduce a \textbf{Camera Predictor} module that estimates the camera parameters directly from the input image features. The predicted viewpoint is then used to supervise the generation of the 3D feature representation, ensuring it remains view-consistent even in the absence of explicit camera inputs at inference time.

To generate the 3D feature volume $\mathcal{V}_{out}$, we define a spatial positional embedding $\mathcal{V}_{in}$ to serve as the query, which is then fed into Cross-Attention to integrate information from $\mathcal{F}_{ref}$, as illustrated in Figure~\ref{fig:arch}b. This mechanism injects information beyond the current viewpoint, including identity, expression, attributes, and other contextual details. To condition the 3D representation on the input camera viewpoint, we introduce a Time-Aware Camera Modulation Block based on an Adaptive Layer Normalization, denoted as ${ModLN}$ (the Modulation Block in Figure~\ref{fig:arch}b)
\begin{equation} \label{eq:modln}
    \begin{gathered}
       \gamma, \; \beta = MLP(\mathcal{C}_{in}), \\
        ModLN(f^{\prime}) = LayerNorm(f) \cdot (1 + \gamma) + \beta + t,
    \end{gathered}
\end{equation}
which is applied between every attention module and feed-forward layer within the Transformer. 

\paragraph{2D Features Sampling and Aggregation.}
Following ~\cite{liu2023syncdreamer} and ~\cite{yang2024consistnet}, the 3D grid features are then warpped into frustum volume features corresponding to the specified camera viewpoint $\mathcal{C}^{i}$ by ray sampling and feature interpolation along each viewing direction
\begin{align}
\mathcal{F}_{frustum}^{i} = Warp(\mathcal{V}_{out}, \; \mathcal{C}^{i}), \; \mathcal{F}_{frustum}^{i} \in \mathbb{R}^{D \times H \times W \times C}.
\end{align}
Since our method operates entirely in the latent space, we do not adopt volumetric rendering as used in NeRF~\cite{mildenhall2021nerf, zhang2022fdnerf}, which fuses depth information based on explicit physical modeling. Instead, we design a \textbf{Depth Aggregation Transformer} ${Aggr}$, whose modules contain cross-depth attentions and spatial attentions (Fig~\ref{fig:arch}b rightmost), to enhance the expressive ability of the resulting 2D features from each novel view. Finally, we apply average pooling along the depth dimension of the frustum volume to produce novel-view features $\mathcal{F}_{out}^{i}$ that match the spatial dimensions of the input feature $\mathcal{F}_{ref}$.
\begin{align}
\mathcal{F}_{out}^{i} = AvgPool(Aggr(\mathcal{F}_{frustum}^{i})), \; \mathcal{F}_{out}^{i} \in \mathbb{R}^{H \times W \times C}
\end{align}
It is important to note that during the second training step, only the parameters of the 3D Feature Construction Model, Depth Aggregation Transformer and the Camera Predictor are updated. The parameters of the image encoder and the SD model remain frozen, as they were already fine-tuned for the image restoration task during the first training step. Modifying these parameters would negatively impact restoration quality. In summary, our approach decouples the target task into two independent objectives during training but fuses into a unified pipeline during inference. This modular design enhances the model's scalability and robustness, while also simplifying the training complexity.

\subsection{Loss Functions} \label{sec:subsection3.3}
In the second training step, we use the same diffusion loss as in Eq \ref{eq:sdloss} but additionally introduce a feature loss $\mathcal{L}_{feat}$ to align the generated novel-view features with their ground truth counterparts. The ground truth features are obtained by first degrading the novel-view ground truth images $\mathcal{I}_{tgt}$ using the same degradation level applied to $\mathcal{I}_{ref}$. These degraded images are then passed through the pretrained image encoder from the first step to extract the ground truth novel-view features $\mathcal{F}_{tgt}$. Since the SD model has already learned how to restore high-resolution images from degraded features, applying the same noise pattern ensures that the generated features and ground truth features share the same distribution. This allows the model to focus solely on novel-view feature generation rather than on restoration, effectively decoupling the two tasks. The feature loss is formulated as below
\begin{equation}
    \begin{gathered}
        \mathcal{F}_{tgt}^{i} = Enc(Degrade(\mathcal{I}_{tgt}^{i}), \; t), \\
        \mathcal{L}_{feat} = \frac{1}{N} \sum_{i=1}^{N}\left(
            \left(\mathcal{F}_{out}^{i} - \mathcal{F}_{tgt}^{i} \right)^2 
            + \lambda \left(1 - \smash{\frac{\mathcal{F}_{out}^{i} \cdot \mathcal{F}_{tgt}^{i}}
            {\lVert \mathcal{F}_{out}^{i} \rVert \lVert \mathcal{F}_{tgt}^{i} \rVert}}\right)
        \right)
    \end{gathered}
\end{equation}
where $N$ is the number of camera views and $\mathcal{L}_{feat}$ comprises two components, MSE Loss and Square Cosine Loss, to jointly constrain the generated features at pixel level and feature space level. Additionally, to ensure that the camera parameters predicted by the Camera Predictor $\mathcal{C}_{in}$ align with the ground truth of the input image, we compute an MSE loss between the predicted and ground-truth camera parameters, which referred to as the camera loss, 
\begin{equation}
\mathcal{L}_{cam} = \left\| C_{\text{in}} - C_{\text{real}} \right\|_2^2
\end{equation}
Consequently, the total loss function in the second training step is defined as
\begin{equation} \label{eq:total}
    \mathcal{L}_{total} = \mathcal{L}_{SD} + \lambda_1 \mathcal{L}_{feat} + \lambda_2 \mathcal{L}_{cam}.
\end{equation}

\section{Experiments}
\subsection{Experimental Settings} \label{sec:subsection4.1}
\paragraph{Datasets.} Similar to the dataset setting in~\cite{gu2024diffportrait3d}, we use a mixed training dataset for novel view synthesis in the second training step, which includes both photo-realistic multi-view dataset NeRSemble~\cite{kirschstein2023nersemble} and synthetic data generated using PanoHead~\cite{an2023panohead}. NeRSemble dataset contains dynamic expression videos of 220 subjects captured from 16 synchronized fixed viewpoints. We sample 3000 sets of multi-view frames from this dataset. To expand the range of viewpoints, increase subject diversity, and enhance background variance, we further augment the training data with 3000 sets of images generated using PanoHead. This augmentation addresses the low diversity inherent in NeRSemble, as it is a lab-controlled dataset with limited variability. For our first training step, we enhance the model's adaptability to multi-view faces by incorporating the above-mentioned datasets. Additionally, we mix in the high-resolution face dataset FFHQ~\cite{karras2019style} to improve the model's generalization ability. During training, all input images are cropped and aligned following the preprocessing strategy used in EG3D. To generate the LQ input images $\mathcal{I}_{ref}$, we adopt the Real-ESRGAN~\cite{wang2021real} pipeline for degradation simulation. For quantitative evaluation, we sample 300 unseen sets of multi-view images from the NeRSemble dataset. To evaluate our model's generalization ability on in-the-wild data, we perform qualitative evaluation on LFW-Test~\cite{huang2008labeled} dataset and quantitative evaluation for single view reconstruction on CelebA-Test~\cite{karras2017progressive} dataset.

\begin{figure}[t]
\centering
\includegraphics[width=\linewidth]{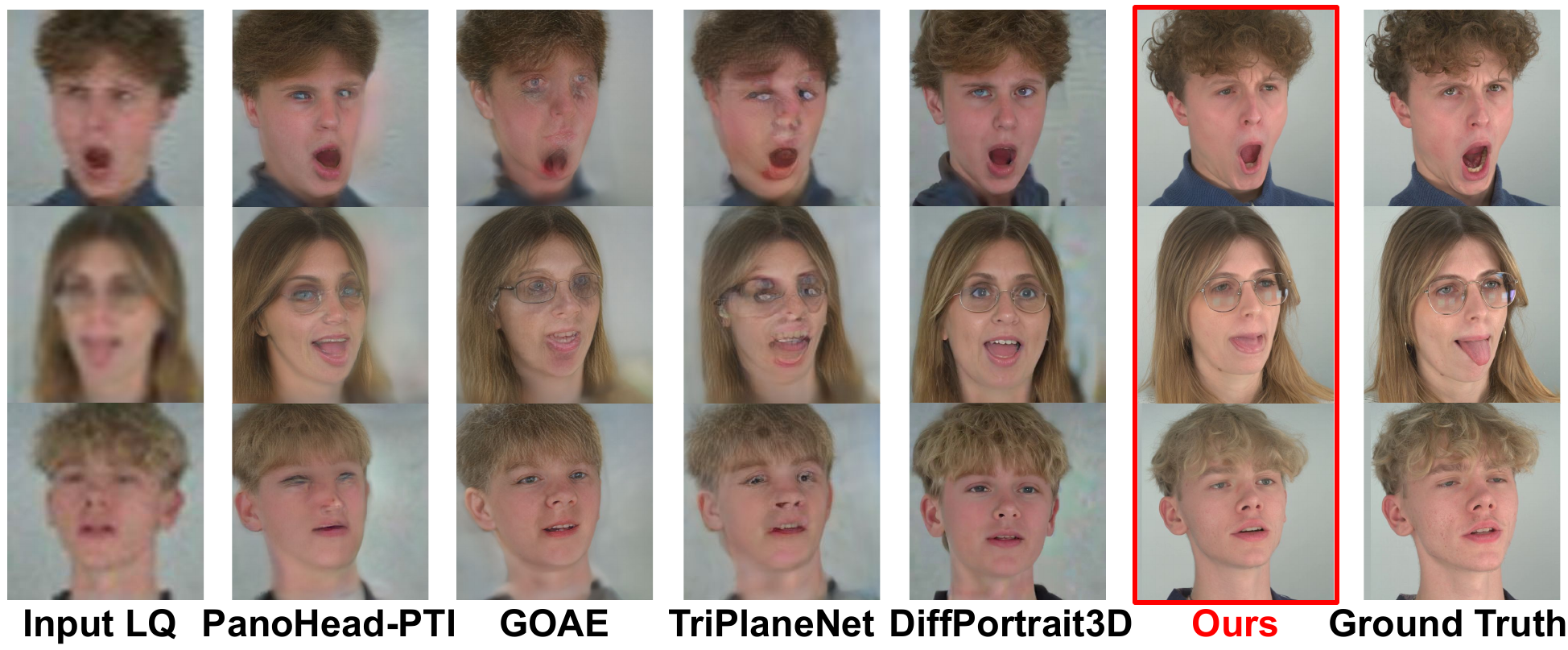}
\caption{Qualitative comparisons on NeRSemble~\cite{kirschstein2023nersemble} dataset. As shown in our results, this end-to-end strategy achieves superior perceptual quality and preserves identity and expression information more effectively than two-stage methods, minimizing the loss of critical facial attributes.}
\label{fig:nersemble}
\end{figure}

\paragraph{Training details.} In the first training step, the image encoder, the cross-attention layers of the SD model, and the LoRA parameters (with a rank of 16) are jointly optimized for 200K interations, with a learning rate of 5e-5 and a batch size of 8. In the second training step, we set the hyper-parameters in Eq \ref{eq:trans} as $\lambda=2.0$, $\lambda_1=10$ and $\lambda_2=0.05$ in Eq \ref{eq:total}. For each input image, we randomly sample 7 different target viewpoints from the remaining views (the total views $N=8$) to supervise the generation of the latent 3D representation. This step is trained with a learning rate of $1e-4$, a batch size of 2 for 300K iterations. During inference, we adopt the DDIM~\cite{song2020denoising} denoising sampler with 50 steps. All training and inference experiments are conducted on a cluster of 8 NVIDIA RTX 6000 Ada GPUs.

\begin{figure*}
\centering
\includegraphics[width=0.96\textwidth]{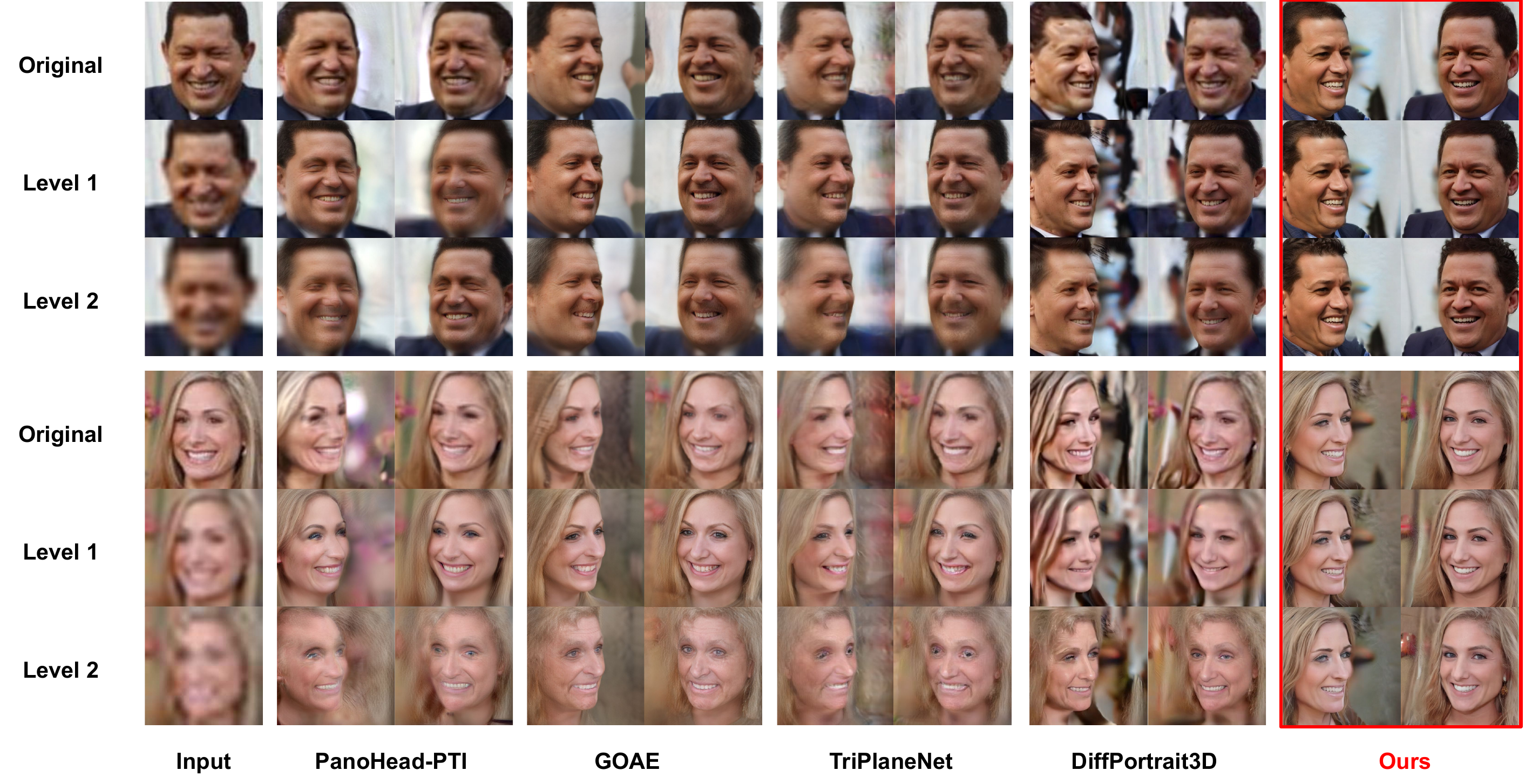}
\caption{Qualitative comparisons on LFW-Test~\cite{huang2008labeled} dataset. Our method produces consistently stable results across varying levels of input degradation. Compared to other approaches, our generated images preserve the most information from the original input and exhibit higher visual realism, even under severe degradation.}
\label{fig:lfw}
\end{figure*}

\subsection{Qualitative Comparisons} \label{sec:subsection4.2}
We conduct comparative experiments with several state-of-the-art novel-view synthesis methods, including PanoHead-PTI~\cite{an2023panohead}, GOAE~\cite{yuan2023make}, TriPlaneNet~\cite{bhattarai2024triplanenet} and DiffPortrait3D~\cite{gu2024diffportrait3d}, and present qualitative comparisons on the NeRSemble~\cite{kirschstein2023nersemble} and LFW-Test~\cite{huang2008labeled} datasets in Figure~\ref{fig:nersemble} and Figure~\ref{fig:lfw}, respectively. Among the compared methods, DiffPortrait3D is based on SD model, while the others are GAN-based. Since these existing methods are not designed to handle degraded facial images, we first apply CodeFormer~\cite{zhou2022towards} for blind face restoration before passing the results to each method for novel view generation. Notably, GAN-based approaches require accurate camera parameters as input to perform novel view synthesis. However, due to the degraded quality of the CodeFormer-restored images, it is often difficult to extract reliable camera parameters from them. To ensure successful generation for these methods, we can only instead use the original high-quality images to extract the input camera parameters. Actually this is impossible in real-world blind restoration, hence highlights a critical limitation of two-stage methods: poor restoration quality makes accurate camera parameter estimation for novel view synthesis intractable. Our single-stage approach avoids this entirely, as it does not rely on external parameter estimation.

\begin{table*}[h!]
    \small
    \centering
    \begin{subtable}[t]{1.5\columnwidth}
    \begin{tabular}{p{3em}>{\centering\arraybackslash}m{6em}>{\centering\arraybackslash}m{4em}>{\centering\arraybackslash}m{5em}>{\centering\arraybackslash}m{5em}>{\centering\arraybackslash}m{5em}}
        \toprule
        &PanoHead-PTI &GOAE &TriplaneNet &DiffPortrait3D &Ours   \\
        \midrule
        SSIM$\uparrow$  &0.73/0.62 &0.73/0.58 &0.75/0.63 &0.68/0.62 &\textbf{0.78/0.63}  \\
        \midrule
        LPIPS$\downarrow$ &0.48/0.50 &0.45/0.53 &0.51/0.51 &0.52/0.49 &\textbf{0.17/0.49} \\
        \midrule
        DISTS$\downarrow$ &0.26/0.28 &0.25/0.27 &0.32/0.29 &0.26/0.27 &\textbf{0.10/0.22}    \\
        \midrule
        FID$\downarrow$ &83.19/65.88 &90.00/71.12 &111.84/63.63 &80.16/67.06 &\textbf{5.67/22.71}  \\
        \midrule
        ID$\uparrow$  &0.32/0.22 &0.27/0.21 &0.30/0.34 &0.29/0.21 &\textbf{0.77/0.46}  \\
        \midrule
        POSE$\downarrow$ &0.0446/- &0.0425/- &0.0456/- &0.0341/- &\textbf{0.0084/-} \\
        \bottomrule
    \end{tabular}
    \caption{}
    \end{subtable}
    \hfill
    \begin{subtable}[t]{0.42\columnwidth}
        \begin{tabular}{>{\centering\arraybackslash}m{2.2em}>{\centering\arraybackslash}m{2.0em}>{\centering\arraybackslash}m{2.0em}}
        \toprule
        & w/o $\mathcal{L}_{feat}$  & w/ $\mathcal{L}_{feat}$ \\
        \midrule
            SSIM$\uparrow$  &0.75 &\textbf{0.78}  \\
            \midrule
            LPIPS$\downarrow$ &0.24 &\textbf{0.17}  \\
            \midrule
            DISTS$\downarrow$ &0.16 &\textbf{0.10}  \\
            \midrule
            FID$\downarrow$ &11.72 &\textbf{5.67}  \\
            \midrule
            ID$\uparrow$  &0.57 &\textbf{0.77}  \\
            \bottomrule
        \end{tabular}   
        \caption{}
    \end{subtable}
    \caption{(a) Quantitative comparisons with other methods on NeRSemble dataset (novel-view synthesis) / CelebA-Test (single-view reconstruction). (b) Ablation study on NeRSemble dataset to assess the effectiveness of the feature loss in our framework. Specifically, we compare the quantitative performance of our model with and without feature loss.
    }
\label{tab:table}
\end{table*}


As shown in Figure~\ref{fig:nersemble}, our method achieves results that are closest to the ground truth in terms of both identity and expression preservation. In contrast, restoration-based pipelines inevitably introduce errors during the blind restoration process, such as identity shift, expression shift, and visual artifacts. When these imperfect restorations are used as inputs for novel view generation, the errors are often amplified, leading to substantial information loss and noticeable degradation in image quality. We further evaluate our method on the LFW-Test dataset to demonstrate its robustness under varying levels of degradation in real-world conditions. We adjust the hyper-parameters of Real-ESRGAN~\cite{wang2021real} to generate degradation level 1 and 2, from weak to strong. As shown in Figure~\ref{fig:lfw}, when the input is a clean image, our method produces results comparable to prior approaches. However, as the degradation level increases, the performance of other methods drops significantly, resulting in poor-quality novel views. While our method may also exhibit some identity and expression shifts, these changes are generally plausible predictions based on the degraded input, and the generated results remain substantially more stable compared to other approaches.

\subsection{Quantitative Comparisons} \label{sec:subsection4.3}
Similar to~\cite{gu2024diffportrait3d}, we evaluate our method on both multi-view generation and single-view reconstruction tasks. For quantitative evaluation, we adopt a comprehensive set of metrics, including LPIPS~\cite{zhang2018unreasonable}, DISTS~\cite{ding2020image}, SSIM~\cite{wang2004image}, ID similarity, FID~\cite{heusel2017gans}, and POSE error. 
The ID similarity is computed by extracting face embeddings from the generated images using the method in~\cite{deng2019arcface}, followed by measuring the cosine similarity with the ground truth embeddings. The POSE error is computed by estimating the facial pose using the approach from~\cite{deng2019accurate} and calculating the Mean Squared Error (MSE) with respect to the estimated pose from ground truth images. For the multi-view generation task, we use our unseen test split of the NeRSemble dataset, where the model generates all the other novel views from a single degraded input image. For the single-view reconstruction task, we use degraded images from the CelebA-Test dataset and aim to reconstruct the original view from the same pose. Competing methods follow a two-stage pipeline, where the degraded image is first restored using CodeFormer~\cite{zhou2022towards} before novel-view synthesis.

As shown in Table~\ref{tab:table}a, our proposed method outperforms all baseline approaches across all evaluation metrics. These results validate that our model can not only faithfully preserve the original content, but also robustly generate novel views, even from heavily degraded inputs.

\subsection{Ablation Studies} \label{sec:subsection4.4}
\begin{figure}[t]
\centering
\includegraphics[width=\linewidth]{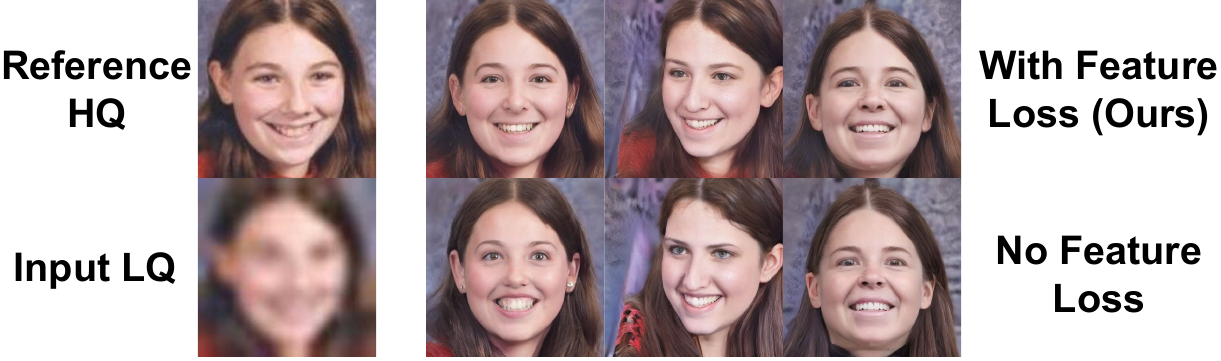}
\caption{Qualitative ablation study on LFW-Test dataset to compare our method with and without feature loss.}
\label{fig:ablation1}
\end{figure}

\paragraph{Without Feature Loss.} 
We conduct ablation studies on both NeRSemble and LFW-Test datasets to evaluate the impact of the feature loss on both quantitative and qualitative performance. In the ablated setting, we remove the feature loss and optimize only the diffusion loss, keeping all other training procedures identical. As shown in Table~\ref{tab:table}b, removing the feature loss results in a significant drop in performance on NeRSemble test set. To further investigate which aspect of performance is most affected, we evaluate the model on the LFW-Test dataset, which contains challenging in-the-wild samples. As shown in Figure~\ref{fig:ablation1}, removing the feature loss leads to severe multi-view inconsistency in the generated outputs. We attribute this issue to the lack of explicit constraints in the latent space during training. Without feature loss, the model is only optimized to produce visually plausible final images, while ignoring the alignment and consistency of novel-view features in latent space. Although the powerful generative capacity of the diffusion model can superficially compensate for misaligned features during training, this shortcut fails to generalize. When tested on unseen data, the unaligned latent features lead to noticeable multi-view inconsistency, severely degrading the reliability of the generated results.

\begin{figure}[t]
\centering
\includegraphics[width=\linewidth]{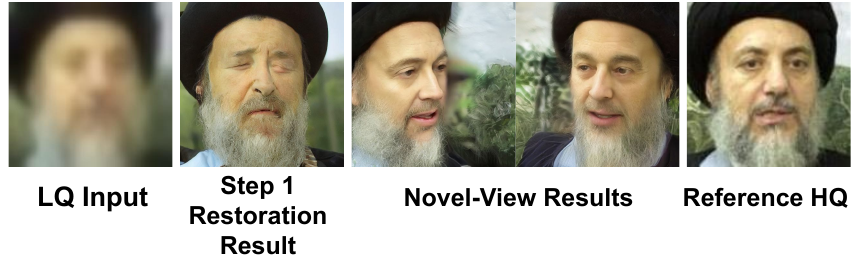}
\caption{Feature correction of our training step 2.}
\label{fig:ablation2}
\end{figure}

\paragraph{Feature Correction in Training Step 2}
We further investigate whether suboptimal image restoration in training step 1 would negatively affect the final novel-view synthesis results. To this end, we use the image restoration model of training step 1 to recover high-resolution images from degraded inputs, and compare these with the final novel-view outputs produced by our whole NVB-Face model. As shown in Figure~\ref{fig:ablation2}, even when the restored images from training step 1 are not visually satisfactory—indicating that the image encoder failed to extract all critical information—our training step 2 is still able to correct the imperfect features and produce more realistic novel-view images. This is made possible by our viewpoint transformation model, which is trained independently in training step 2. As a result, it can generate robust novel-view features even when the input features are partially flawed. As discussed earlier, traditional two-stage methods tend to amplify errors during novel-view synthesis when the restoration quality is suboptimal. In contrast, our single-stage approach is able to mitigate such error propagation, resulting in more stable and reliable outputs.

\section{Conclusion}
In this paper, we propose NVB-Face, the first framework capable of directly generating novel views from a single blind face image. NVB-Face decouples the overall task into two carefully designed and independently optimized steps, while introducing tailored loss functions that seamlessly integrate the two components. In contrast to traditional two-stage pipelines that perform face restoration followed by novel-view synthesis, our end-to-end approach significantly reduces error accumulation and better preserves identity and other essential image attributes. Furthermore, by constructing novel-view features in a 3D latent space, our method effectively enforces multi-view consistency across generated images. Extensive experiments demonstrate that our NVB-Face reliably synthesizes high-quality and consistent novel views for arbitrary in-the-wild blind face images.

\section{Acknowledgments}
This research is based upon work supported in part by the Office of the Director of National Intelligence (ODNI), Intelligence Advanced Research Projects Activity (IARPA), via [2022-21102100003]. 
The views and conclusions contained herein are those of the authors and should not be interpreted as necessarily representing the official policies, either expressed or implied, of IARPA, or the U.S. Government. 
The U.S. Government is authorized to reproduce and distribute reprints for governmental purposes notwithstanding any copyright annotation therein. 


\bibliography{aaai2026}

@String(TOG= {ACM Trans. Graph.})

@String(ICLR = {Int. Conf. Learn. Represent.})

@String(TOG   = {ACM TOG})

@String(ICLR  = {ICLR})

@article{wang2004image,
  title={Image quality assessment: from error visibility to structural similarity},
  author={Wang, Zhou and Bovik, Alan C and Sheikh, Hamid R and Simoncelli, Eero P},
  journal={IEEE transactions on image processing},
  volume={13},
  number={4},
  pages={600--612},
  year={2004},
  publisher={IEEE}
}

@inproceedings{huang2008labeled,
  title={Labeled faces in the wild: A database forstudying face recognition in unconstrained environments},
  author={Huang, Gary B and Mattar, Marwan and Berg, Tamara and Learned-Miller, Eric},
  booktitle={Workshop on faces in'Real-Life'Images: detection, alignment, and recognition},
  year={2008}
}

@inproceedings{paysan20093d,
  title={A 3D face model for pose and illumination invariant face recognition},
  author={Paysan, Pascal and Knothe, Reinhard and Amberg, Brian and Romdhani, Sami and Vetter, Thomas},
  booktitle={2009 sixth IEEE international conference on advanced video and signal based surveillance},
  pages={296--301},
  year={2009},
  organization={Ieee}
}

@article{karras2017progressive,
  title={Progressive growing of gans for improved quality, stability, and variation},
  author={Karras, Tero and Aila, Timo and Laine, Samuli and Lehtinen, Jaakko},
  journal={arXiv preprint arXiv:1710.10196},
  year={2017}
}

@article{heusel2017gans,
  title={Gans trained by a two time-scale update rule converge to a local nash equilibrium},
  author={Heusel, Martin and Ramsauer, Hubert and Unterthiner, Thomas and Nessler, Bernhard and Hochreiter, Sepp},
  journal={Advances in neural information processing systems},
  volume={30},
  year={2017}
}

@inproceedings{zhang2018unreasonable,
  title={The unreasonable effectiveness of deep features as a perceptual metric},
  author={Zhang, Richard and Isola, Phillip and Efros, Alexei A and Shechtman, Eli and Wang, Oliver},
  booktitle={Proceedings of the IEEE conference on computer vision and pattern recognition},
  pages={586--595},
  year={2018}
}

@inproceedings{deng2019accurate,
  title={Accurate 3d face reconstruction with weakly-supervised learning: From single image to image set},
  author={Deng, Yu and Yang, Jiaolong and Xu, Sicheng and Chen, Dong and Jia, Yunde and Tong, Xin},
  booktitle={Proceedings of the IEEE/CVF conference on computer vision and pattern recognition workshops},
  pages={0--0},
  year={2019}
}

@inproceedings{deng2019arcface,
  title={Arcface: Additive angular margin loss for deep face recognition},
  author={Deng, Jiankang and Guo, Jia and Xue, Niannan and Zafeiriou, Stefanos},
  booktitle={Proceedings of the IEEE/CVF conference on computer vision and pattern recognition},
  pages={4690--4699},
  year={2019}
}

@inproceedings{wu2019mvf,
  title={Mvf-net: Multi-view 3d face morphable model regression},
  author={Wu, Fanzi and Bao, Linchao and Chen, Yajing and Ling, Yonggen and Song, Yibing and Li, Songnan and Ngan, King Ngi and Liu, Wei},
  booktitle={Proceedings of the IEEE/CVF conference on computer vision and pattern recognition},
  pages={959--968},
  year={2019}
}

@inproceedings{karras2019style,
  title={A style-based generator architecture for generative adversarial networks},
  author={Karras, Tero and Laine, Samuli and Aila, Timo},
  booktitle={Proceedings of the IEEE/CVF conference on computer vision and pattern recognition},
  pages={4401--4410},
  year={2019}
}

@article{ding2020image,
  title={Image quality assessment: Unifying structure and texture similarity},
  author={Ding, Keyan and Ma, Kede and Wang, Shiqi and Simoncelli, Eero P},
  journal={IEEE transactions on pattern analysis and machine intelligence},
  volume={44},
  number={5},
  pages={2567--2581},
  year={2020},
  publisher={IEEE}
}

@article{song2020denoising,
  title={Denoising diffusion implicit models},
  author={Song, Jiaming and Meng, Chenlin and Ermon, Stefano},
  journal={arXiv preprint arXiv:2010.02502},
  year={2020}
}

@inproceedings{yang2021gan,
  title={Gan prior embedded network for blind face restoration in the wild},
  author={Yang, Tao and Ren, Peiran and Xie, Xuansong and Zhang, Lei},
  booktitle={Proceedings of the IEEE/CVF conference on computer vision and pattern recognition},
  pages={672--681},
  year={2021}
}

@inproceedings{wang2021towards,
  title={Towards real-world blind face restoration with generative facial prior},
  author={Wang, Xintao and Li, Yu and Zhang, Honglun and Shan, Ying},
  booktitle={Proceedings of the IEEE/CVF conference on computer vision and pattern recognition},
  pages={9168--9178},
  year={2021}
}

@inproceedings{esser2021taming,
  title={Taming transformers for high-resolution image synthesis},
  author={Esser, Patrick and Rombach, Robin and Ommer, Bjorn},
  booktitle={Proceedings of the IEEE/CVF conference on computer vision and pattern recognition},
  pages={12873--12883},
  year={2021}
}

@inproceedings{dib2021towards,
  title={Towards high fidelity monocular face reconstruction with rich reflectance using self-supervised learning and ray tracing},
  author={Dib, Abdallah and Thebault, Cedric and Ahn, Junghyun and Gosselin, Philippe-Henri and Theobalt, Christian and Chevallier, Louis},
  booktitle={Proceedings of the IEEE/CVF International Conference on Computer Vision},
  pages={12819--12829},
  year={2021}
}

@inproceedings{chan2021pi,
  title={pi-gan: Periodic implicit generative adversarial networks for 3d-aware image synthesis},
  author={Chan, Eric R and Monteiro, Marco and Kellnhofer, Petr and Wu, Jiajun and Wetzstein, Gordon},
  booktitle={Proceedings of the IEEE/CVF conference on computer vision and pattern recognition},
  pages={5799--5809},
  year={2021}
}

@inproceedings{wang2021real,
  title={Real-esrgan: Training real-world blind super-resolution with pure synthetic data},
  author={Wang, Xintao and Xie, Liangbin and Dong, Chao and Shan, Ying},
  booktitle={Proceedings of the IEEE/CVF international conference on computer vision},
  pages={1905--1914},
  year={2021}
}

@inproceedings{radford2021learning,
  title={Learning transferable visual models from natural language supervision},
  author={Radford, Alec and Kim, Jong Wook and Hallacy, Chris and Ramesh, Aditya and Goh, Gabriel and Agarwal, Sandhini and Sastry, Girish and Askell, Amanda and Mishkin, Pamela and Clark, Jack and others},
  booktitle={International conference on machine learning},
  pages={8748--8763},
  year={2021},
  organization={PmLR}
}

@article{gu2021stylenerf,
  title={Stylenerf: A style-based 3d-aware generator for high-resolution image synthesis},
  author={Gu, Jiatao and Liu, Lingjie and Wang, Peng and Theobalt, Christian},
  journal={arXiv preprint arXiv:2110.08985},
  year={2021}
}

@article{mildenhall2021nerf,
  title={Nerf: Representing scenes as neural radiance fields for view synthesis},
  author={Mildenhall, Ben and Srinivasan, Pratul P and Tancik, Matthew and Barron, Jonathan T and Ramamoorthi, Ravi and Ng, Ren},
  journal={Communications of the ACM},
  volume={65},
  number={1},
  pages={99--106},
  year={2021},
  publisher={ACM New York, NY, USA}
}

@inproceedings{wang2022restoreformer,
  title={Restoreformer: High-quality blind face restoration from undegraded key-value pairs},
  author={Wang, Zhouxia and Zhang, Jiawei and Chen, Runjian and Wang, Wenping and Luo, Ping},
  booktitle={Proceedings of the IEEE/CVF conference on computer vision and pattern recognition},
  pages={17512--17521},
  year={2022}
}

@article{zhou2022towards,
  title={Towards robust blind face restoration with codebook lookup transformer},
  author={Zhou, Shangchen and Chan, Kelvin and Li, Chongyi and Loy, Chen Change},
  journal={Advances in Neural Information Processing Systems},
  volume={35},
  pages={30599--30611},
  year={2022}
}

@inproceedings{zhuang2022mofanerf,
  title={Mofanerf: Morphable facial neural radiance field},
  author={Zhuang, Yiyu and Zhu, Hao and Sun, Xusen and Cao, Xun},
  booktitle={European conference on computer vision},
  pages={268--285},
  year={2022},
  organization={Springer}
}

@inproceedings{zhang2022fdnerf,
  title={Fdnerf: Few-shot dynamic neural radiance fields for face reconstruction and expression editing},
  author={Zhang, Jingbo and Li, Xiaoyu and Wan, Ziyu and Wang, Can and Liao, Jing},
  booktitle={SIGGRAPH Asia 2022 Conference Papers},
  pages={1--9},
  year={2022}
}

@inproceedings{rombach2022high,
  title={High-resolution image synthesis with latent diffusion models},
  author={Rombach, Robin and Blattmann, Andreas and Lorenz, Dominik and Esser, Patrick and Ommer, Bj{\"o}rn},
  booktitle={Proceedings of the IEEE/CVF conference on computer vision and pattern recognition},
  pages={10684--10695},
  year={2022}
}

@inproceedings{chan2022efficient,
  title={Efficient geometry-aware 3d generative adversarial networks},
  author={Chan, Eric R and Lin, Connor Z and Chan, Matthew A and Nagano, Koki and Pan, Boxiao and De Mello, Shalini and Gallo, Orazio and Guibas, Leonidas J and Tremblay, Jonathan and Khamis, Sameh and others},
  booktitle={Proceedings of the IEEE/CVF conference on computer vision and pattern recognition},
  pages={16123--16133},
  year={2022}
}

@article{hu2022lora,
  title={Lora: Low-rank adaptation of large language models.},
  author={Hu, Edward J and Shen, Yelong and Wallis, Phillip and Allen-Zhu, Zeyuan and Li, Yuanzhi and Wang, Shean and Wang, Lu and Chen, Weizhu and others},
  journal={ICLR},
  volume={1},
  number={2},
  pages={3},
  year={2022}
}

@article{luo2023lcm,
  title={Lcm-lora: A universal stable-diffusion acceleration module},
  author={Luo, Simian and Tan, Yiqin and Patil, Suraj and Gu, Daniel and von Platen, Patrick and Passos, Apolin{\'a}rio and Huang, Longbo and Li, Jian and Zhao, Hang},
  journal={arXiv preprint arXiv:2311.05556},
  year={2023}
}

@article{wang2023restoreformer++,
  title={RestoreFormer++: Towards real-world blind face restoration from undegraded key-value pairs},
  author={Wang, Zhouxia and Zhang, Jiawei and Chen, Tianshui and Wang, Wenping and Luo, Ping},
  journal={IEEE Transactions on Pattern Analysis and Machine Intelligence},
  year={2023},
  publisher={IEEE}
}

@inproceedings{an2023panohead,
  title={Panohead: Geometry-aware 3d full-head synthesis in 360deg},
  author={An, Sizhe and Xu, Hongyi and Shi, Yichun and Song, Guoxian and Ogras, Umit Y and Luo, Linjie},
  booktitle={Proceedings of the IEEE/CVF conference on computer vision and pattern recognition},
  pages={20950--20959},
  year={2023}
}

@article{liu2023syncdreamer,
  title={Syncdreamer: Generating multiview-consistent images from a single-view image},
  author={Liu, Yuan and Lin, Cheng and Zeng, Zijiao and Long, Xiaoxiao and Liu, Lingjie and Komura, Taku and Wang, Wenping},
  journal={arXiv preprint arXiv:2309.03453},
  year={2023}
}

@inproceedings{yin20233d,
  title={3d gan inversion with facial symmetry prior},
  author={Yin, Fei and Zhang, Yong and Wang, Xuan and Wang, Tengfei and Li, Xiaoyu and Gong, Yuan and Fan, Yanbo and Cun, Xiaodong and Shan, Ying and Oztireli, Cengiz and others},
  booktitle={Proceedings of the IEEE/CVF Conference on Computer Vision and Pattern Recognition},
  pages={342--351},
  year={2023}
}

@inproceedings{yuan2023make,
  title={Make encoder great again in 3d gan inversion through geometry and occlusion-aware encoding},
  author={Yuan, Ziyang and Zhu, Yiming and Li, Yu and Liu, Hongyu and Yuan, Chun},
  booktitle={Proceedings of the IEEE/CVF International Conference on Computer Vision},
  pages={2437--2447},
  year={2023}
}

@article{trevithick2023real,
  title={Real-time radiance fields for single-image portrait view synthesis},
  author={Trevithick, Alex and Chan, Matthew and Stengel, Michael and Chan, Eric and Liu, Chao and Yu, Zhiding and Khamis, Sameh and Chandraker, Manmohan and Ramamoorthi, Ravi and Nagano, Koki},
  journal={ACM Transactions on Graphics (TOG)},
  volume={42},
  number={4},
  pages={1--15},
  year={2023},
  publisher={ACM New York, NY, USA}
}

@inproceedings{zhang2023adding,
  title={Adding conditional control to text-to-image diffusion models},
  author={Zhang, Lvmin and Rao, Anyi and Agrawala, Maneesh},
  booktitle={Proceedings of the IEEE/CVF International Conference on Computer Vision},
  pages={3836--3847},
  year={2023}
}

@inproceedings{wang2023dr2,
  title={Dr2: Diffusion-based robust degradation remover for blind face restoration},
  author={Wang, Zhixin and Zhang, Ziying and Zhang, Xiaoyun and Zheng, Huangjie and Zhou, Mingyuan and Zhang, Ya and Wang, Yanfeng},
  booktitle={Proceedings of the IEEE/CVF Conference on Computer Vision and Pattern Recognition},
  pages={1704--1713},
  year={2023}
}

@article{yue2023resshift,
  title={Resshift: Efficient diffusion model for image super-resolution by residual shifting},
  author={Yue, Zongsheng and Wang, Jianyi and Loy, Chen Change},
  journal={Advances in Neural Information Processing Systems},
  volume={36},
  pages={13294--13307},
  year={2023}
}

@article{kirschstein2023nersemble,
  title={Nersemble: Multi-view radiance field reconstruction of human heads},
  author={Kirschstein, Tobias and Qian, Shenhan and Giebenhain, Simon and Walter, Tim and Nie{\ss}ner, Matthias},
  journal={ACM Transactions on Graphics (TOG)},
  volume={42},
  number={4},
  pages={1--14},
  year={2023},
  publisher={ACM New York, NY, USA}
}

@inproceedings{peebles2023scalable,
  title={Scalable diffusion models with transformers},
  author={Peebles, William and Xie, Saining},
  booktitle={Proceedings of the IEEE/CVF international conference on computer vision},
  pages={4195--4205},
  year={2023}
}

@article{hong2023lrm,
  title={Lrm: Large reconstruction model for single image to 3d},
  author={Hong, Yicong and Zhang, Kai and Gu, Jiuxiang and Bi, Sai and Zhou, Yang and Liu, Difan and Liu, Feng and Sunkavalli, Kalyan and Bui, Trung and Tan, Hao},
  journal={arXiv preprint arXiv:2311.04400},
  year={2023}
}

@article{wang2024exploiting,
  title={Exploiting diffusion prior for real-world image super-resolution},
  author={Wang, Jianyi and Yue, Zongsheng and Zhou, Shangchen and Chan, Kelvin CK and Loy, Chen Change},
  journal={International Journal of Computer Vision},
  volume={132},
  number={12},
  pages={5929--5949},
  year={2024},
  publisher={Springer}
}

@inproceedings{yang2024consistnet,
  title={Consistnet: Enforcing 3d consistency for multi-view images diffusion},
  author={Yang, Jiayu and Cheng, Ziang and Duan, Yunfei and Ji, Pan and Li, Hongdong},
  booktitle={Proceedings of the IEEE/CVF Conference on Computer Vision and Pattern Recognition},
  pages={7079--7088},
  year={2024}
}

@inproceedings{lin2024diffbir,
  title={Diffbir: Toward blind image restoration with generative diffusion prior},
  author={Lin, Xinqi and He, Jingwen and Chen, Ziyan and Lyu, Zhaoyang and Dai, Bo and Yu, Fanghua and Qiao, Yu and Ouyang, Wanli and Dong, Chao},
  booktitle={European Conference on Computer Vision},
  pages={430--448},
  year={2024},
  organization={Springer}
}

@article{yue2024difface,
  title={Difface: Blind face restoration with diffused error contraction},
  author={Yue, Zongsheng and Loy, Chen Change},
  journal={IEEE Transactions on Pattern Analysis and Machine Intelligence},
  year={2024},
  publisher={IEEE}
}

@article{sun2024improving,
  title={Improving the stability of diffusion models for content consistent super-resolution},
  author={Sun, Lingchen and Wu, Rongyuan and Zhang, Zhengqiang and Yong, Hongwei and Zhang, Lei},
  journal={CoRR},
  year={2024}
}

@article{wang2024osdface,
  title={OSDFace: One-Step Diffusion Model for Face Restoration},
  author={Wang, Jingkai and Gong, Jue and Zhang, Lin and Chen, Zheng and Liu, Xing and Gu, Hong and Liu, Yutong and Zhang, Yulun and Yang, Xiaokang},
  journal={arXiv preprint arXiv:2411.17163},
  year={2024}
}

@inproceedings{gu2024diffportrait3d,
  title={DiffPortrait3D: Controllable Diffusion for Zero-Shot Portrait View Synthesis},
  author={Gu, Yuming and Xu, Hongyi and Xie, You and Song, Guoxian and Shi, Yichun and Chang, Di and Yang, Jing and Luo, Linjie},
  booktitle={Proceedings of the IEEE/CVF Conference on Computer Vision and Pattern Recognition},
  pages={10456--10465},
  year={2024}
}

@inproceedings{papantoniou2024arc2face,
  title={Arc2face: A foundation model for id-consistent human faces},
  author={Papantoniou, Foivos Paraperas and Lattas, Alexandros and Moschoglou, Stylianos and Deng, Jiankang and Kainz, Bernhard and Zafeiriou, Stefanos},
  booktitle={European Conference on Computer Vision},
  pages={241--261},
  year={2024},
  organization={Springer}
}

@inproceedings{bhattarai2024triplanenet,
  title={Triplanenet: An encoder for eg3d inversion},
  author={Bhattarai, Ananta R and Nie{\ss}ner, Matthias and Sevastopolsky, Artem},
  booktitle={Proceedings of the IEEE/CVF Winter Conference on Applications of Computer Vision},
  pages={3055--3065},
  year={2024}
}

@article{wu2025one,
  title={One-step effective diffusion network for real-world image super-resolution},
  author={Wu, Rongyuan and Sun, Lingchen and Ma, Zhiyuan and Zhang, Lei},
  journal={Advances in Neural Information Processing Systems},
  volume={37},
  pages={92529--92553},
  year={2025}
}

@article{gu2025diffportrait360,
  title={DiffPortrait360: Consistent Portrait Diffusion for 360 View Synthesis},
  author={Gu, Yuming and Tran, Phong and Zheng, Yujian and Xu, Hongyi and Li, Heyuan and Karmanov, Adilbek and Li, Hao},
  journal={arXiv preprint arXiv:2503.15667},
  year={2025}
}
\newpage
\appendix

\section{Details of Our Architecture Design}

\begin{figure}[h!]
\centering
\includegraphics[width=\linewidth]{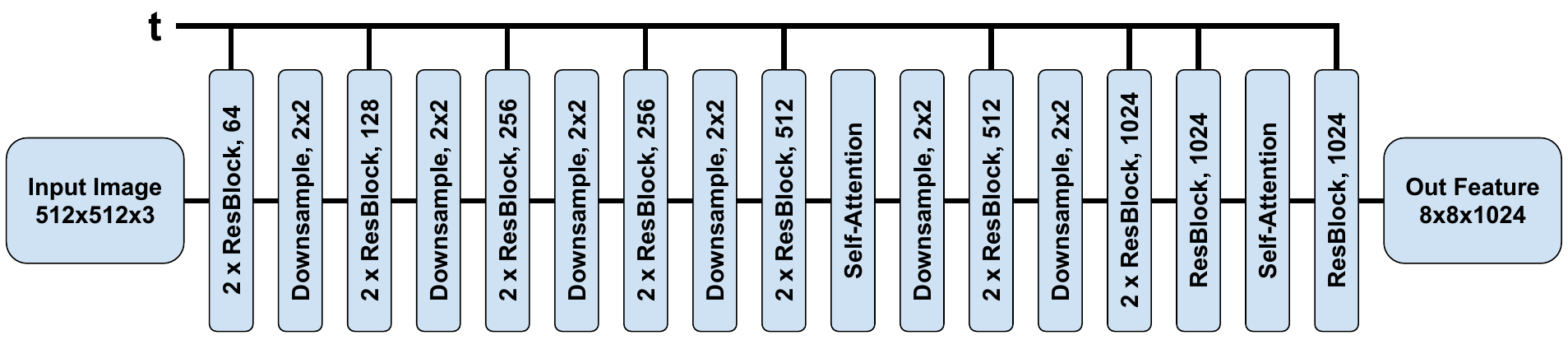}
\caption{Architecture of our Time-aware Image Encoder}
\label{fig:encoder}
\end{figure}

Our image encoder follows the design of ~\cite{esser2021taming,zhou2022towards} (see Figure~\ref{fig:encoder}) and takes an input image of size $512\times512\times3$, producing an output feature map of size $8\times8\times1024$. These extracted features are then fed into either the SD model (in training step 1) via cross-attention, or into the viewpoint transformation model (in training step 2). To ensure alignment with the SD model’s internal representations, we insert the the same time step used by the diffusion process into each ResBlock, following the strategy in~\cite{rombach2022high}. This design ensures temporal and spatial consistency between the encoder output and the diffusion model’s denoising trajectory.

In our training step 2, to implement the 3D-aware viewpoint transformation model efficiently, we apply a memory-efficient design. Specifically, we set the spatial positional embedding $\mathcal{V}_{in}$ to a size of 8×8×8×1024, and the output 3D feature volume $\mathcal{V}_{out}$ maintains the same resolution of 8×8×8×1024. Following the method proposed in ~\cite{liu2023syncdreamer, yang2024consistnet}, we warp this 3D feature volume into 2D feature frustums $\mathcal{F}_{frustum}^{i}$ corresponding to the target camera viewpoints $\mathcal{C}_{i}$, with a depth dimension of 12. Finally, we apply depth aggregation to fuse the depth-wise features into compact 2D representations suitable for novel-view synthesis.

\section{Performance of Blind Face Restoration}
We evaluate our blind face restoration (BFR) pipeline in training step 1 by comparing it with CodeFormer~\cite{zhou2022towards} (GAN-based) and DiffBIR~\cite{lin2024diffbir} (Diffusion-based) on the CelebA-Test dataset. The results are reported in Table~\ref{tab:bfr}. As shown in the table, our BFR module achieves comparable performance to these classic methods, but does not exhibit a significant advantage. This is expected, as our primary goal is not to optimize BFR itself, but rather to integrate BFR and novel-view synthesis into a unified framework. We did not introduce any specialized designs or enhancements specifically for the restoration task. Nonetheless, this comparison further validates the superiority of our single-stage pipeline: even when our BFR performance is on par with other methods, our final novel-view synthesis results are significantly better than those produced by two-stage pipelines based on those methods. This confirms that our approach effectively mitigates error propagation, while traditional two-stage methods tend to amplify restoration errors in the subsequent synthesis stage.

\begin{figure}[t]
\centering
\includegraphics[width=\linewidth]{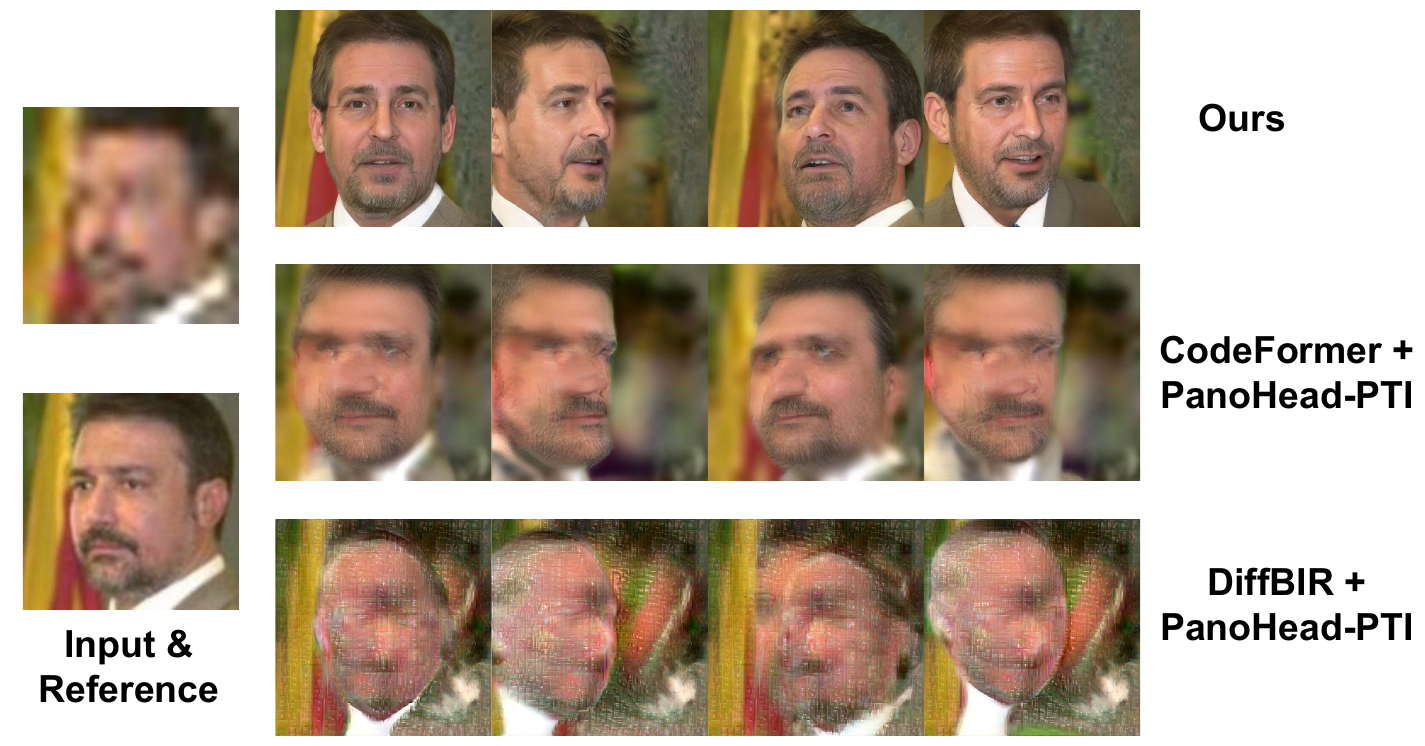}
\caption{Qualitative comparison among CodeFormer-based two-stage method, DiffBIR-based method and ours on LFW-Test dataset}
\label{fig:diffcomp}
\end{figure}

\begin{table*}[t!]
  \centering
  \footnotesize
  \begin{tabular}{p{3.5em}>{\centering\arraybackslash}m{3.5em}>{\centering\arraybackslash}m{2.7em}>{\centering\arraybackslash}m{2.7em}>{\centering\arraybackslash}m{2.7em}>{\centering\arraybackslash}m{2.5em}>{\centering\arraybackslash}m{2.9em}>{\centering\arraybackslash}m{2.8em}>{\centering\arraybackslash}m{3.5em}>{\centering\arraybackslash}m{4.0em}}
    \toprule
        &PSNR$\uparrow$ &SSIM$\uparrow$ &LPIPS$\downarrow$ &DISTS$\downarrow$ &FID$\downarrow$ &MUSIQ$\uparrow$ &NIQE$\downarrow$ &C-IQA$\uparrow$ &M-IQA$\uparrow$   \\
    \midrule
    CodeFormer  &\underline{20.98} &\textbf{0.6174} &\underline{0.4750} &0.2636 &61.44 &66.06 &\underline{5.6992} &0.5608 &0.4775   \\
    \midrule
    DiffBIR  &\textbf{22.00} &\underline{0.6000} &\textbf{0.4631} &\underline{0.2340} &\underline{32.54} &\textbf{75.61} &6.2855 &\textbf{0.7895} &\underline{0.6256}  \\
    \midrule
    Ours    &20.05 &0.5574 &0.4969 &\textbf{0.2133} &\textbf{18.43} &\underline{74.71} &\textbf{5.2747} &\underline{0.7431} &\textbf{0.6536} \\
    \bottomrule
  \end{tabular}
  \caption{Comparison with other classical blind face restoration (BFR) methods. Bold indicates the best result, and underlined indicates the second-best.}
\label{tab:bfr}
\end{table*}

\begin{table*}[t!]
  \centering
  \footnotesize
  \begin{tabular}{llllllll}
    \toprule
        &SSIM$\uparrow$ &LPIPS$\downarrow$ &DISTS$\downarrow$ &FID$\downarrow$ &ID$\uparrow$ &POSE$\downarrow$    \\
    \midrule
    PanoHead-PTI  &0.63(\underline{0.73}) &0.54(\underline{0.48}) &0.27(\underline{0.26}) &85.31(\underline{83.19}) &0.30(\underline{0.32}) &0.0484(\underline{0.0446})   \\
    \midrule
    GOAE &0.62(\underline{0.73}) &0.50(\underline{0.45}) &0.27(\underline{0.25}) &108.65(\underline{90.00}) &\underline{0.27}(\underline{0.27}) &\underline{0.0392}(0.0425)    \\
    \midrule
    TriplaneNet   &0.70(\underline{0.75}) &0.52(\underline{0.51}) &\underline{0.29}(0.32) &136.21(\underline{111.84}) &0.29(\underline{0.30}) &0.0484(\underline{0.0456})  \\
    \midrule
    Ours    &\textbf{0.78} &\textbf{0.17} &\textbf{0.10} &\textbf{5.67} &\textbf{0.77} &\textbf{0.0084} \\
    \bottomrule
  \end{tabular}
  \caption{Quantitative comparisons with DiffBIR-based two-stage methods on NeRSemble dataset. Values in parentheses indicate results based on the CodeFormer restoration. Underlined values represent the better result between the CodeFormer-based and DiffBIR-based variants.}
\label{tab:diffbir}
\end{table*}

\section{Comparison with DiffBIR-based Novel-view Synthesis Methods}
We further compare our method against two-stage pipelines based on DiffBIR~\cite{lin2024diffbir}. As shown in Table~\ref{tab:diffbir}, even when using the diffusion-based DiffBIR model for the first-stage restoration instead of CodeFormer, our approach still achieves superior performance. Moreover, by comparing the results between the CodeFormer-based and DiffBIR-based variants in the table, it is evident that DiffBIR performs worse than CodeFormer in most cases. This is further supported by the qualitative results in Figure~\ref{fig:diffcomp}, where using DiffBIR for the first stage leads to inferior visual quality, especially for inputs with more severe degradation. We speculate that this is due to the fact that DiffBIR was not trained on images with higher levels of degradation, limiting its robustness in such scenarios. Therefore, in all our two-stage baseline comparisons throughout the paper, we consistently use CodeFormer as the restoration module in the first stage to ensure fair and competitive results.

\section{More Results}
\begin{figure*}[p]
\centering
\includegraphics[height=0.95\textheight]{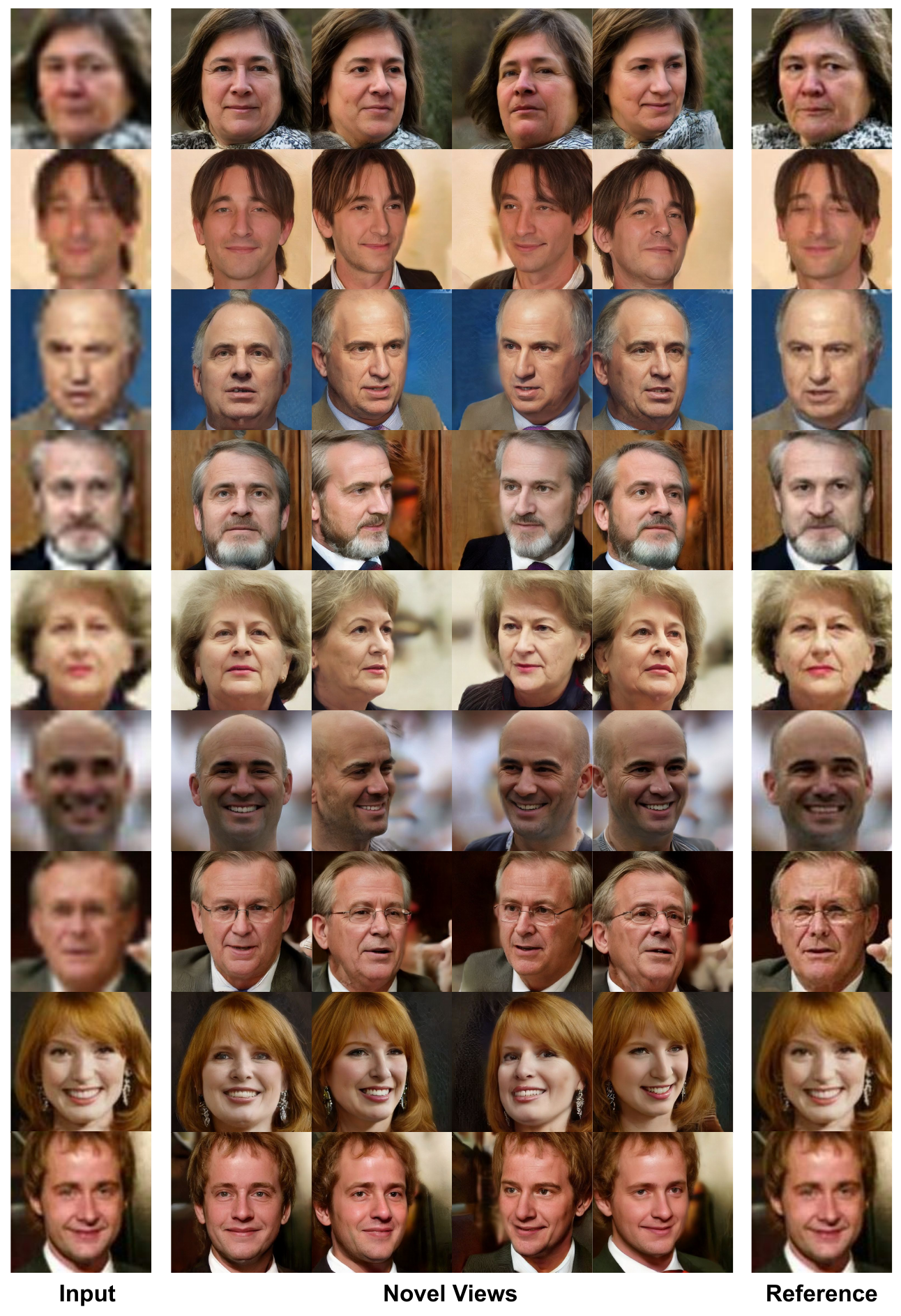}
\caption{More novel-view results from easy cases}
\label{fig:easy}
\end{figure*}

\begin{figure*}[p]
\centering
\includegraphics[height=0.95\textheight]{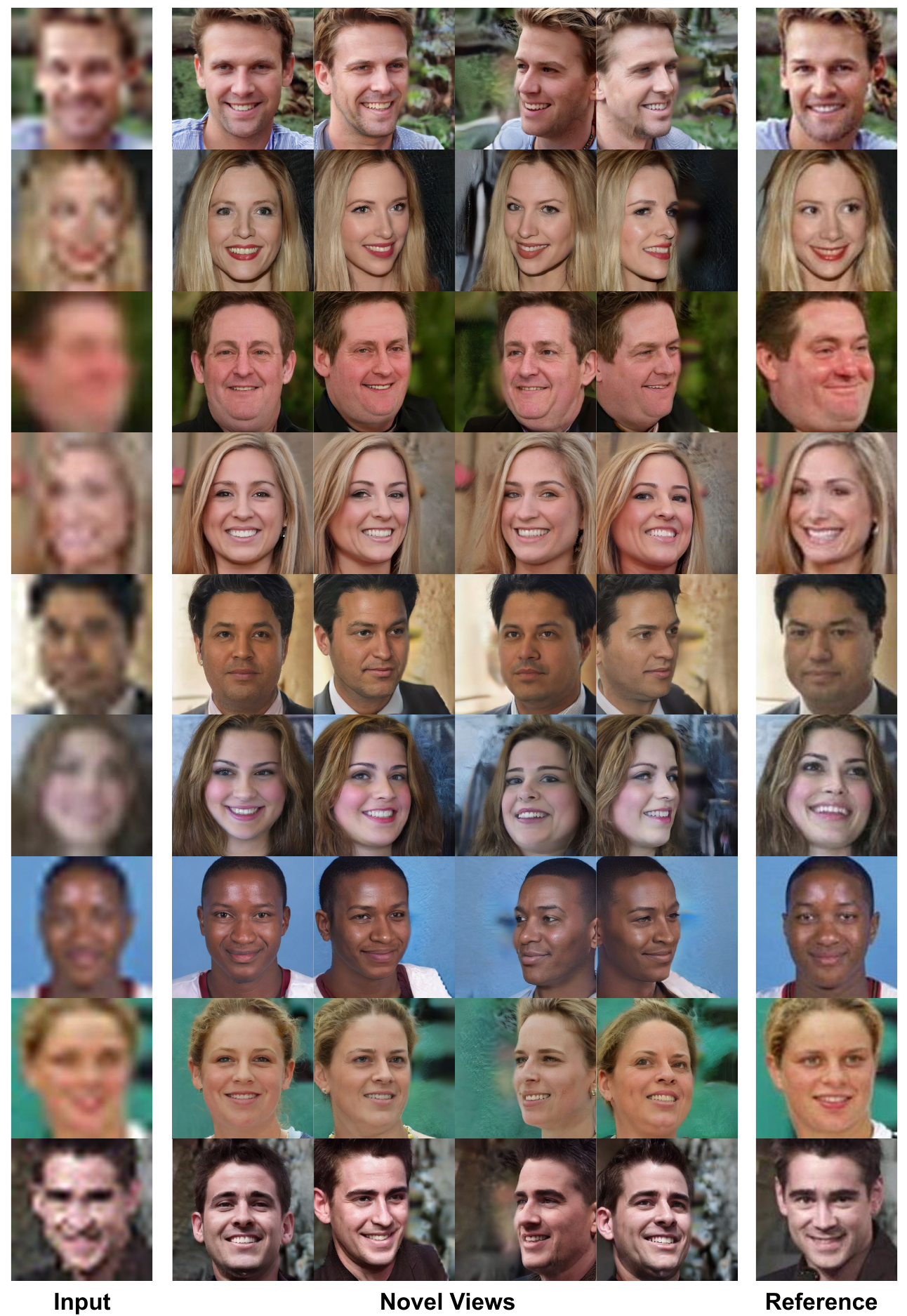}
\caption{More novel-view results from hard cases}
\label{fig:hard}
\end{figure*}

We present additional results in Figure~\ref{fig:easy} and Figure~\ref{fig:hard}, where all input images are sampled from the LFW~\cite{huang2008labeled} dataset. In Figure~\ref{fig:easy}, we showcase cases where the degradation is relatively mild, and in some cases (e.g., the last two rows), the input is the original clean image. Our method consistently produces high-fidelity novel-view images with strong identity preservation and multi-view consistency, performing on par with existing state-of-the-art novel view synthesis methods. In Figure~\ref{fig:hard}, we demonstrate our model’s performance under extremely degraded conditions. Even in such challenging scenarios, our method is able to stably generate plausible and realistic novel-view images. While some identity and expression shifts can be observed, we consider such deviations acceptable given the severe information loss in the input. These results further confirm that, compared to two-stage approaches, our single-stage pipeline exhibits significantly greater robustness under challenging real-world degradations.

\section{Limitations and Future Work}
As discussed in Appendix B, while the performance of our image restoration module is comparable to other blind face restoration (BFR) methods, it does not deliver particularly impressive results. This reinforces the effectiveness of our training step 2, but we acknowledge that better restoration quality in our training step 1 can lead to higher-quality novel-view generation. In future work, we plan to further optimize our training step 1 pipeline, aiming to improve restoration performance while ensuring seamless integration with the novel view synthesis task.

\begin{figure}[h]
\centering
\includegraphics[width=\linewidth]{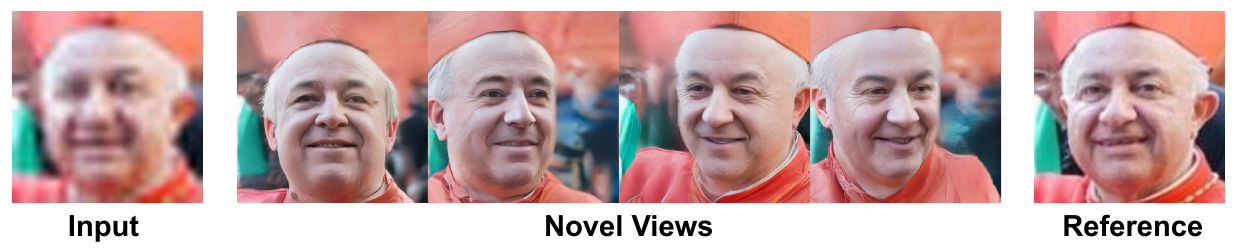}
\caption{Limitation of our model}
\label{fig:limit}
\end{figure}

Moreover, although our model is capable of faithfully generating high-quality and view-consistent facial images, it shows limited generative capability for non-facial content, such as clothing and backgrounds. As shown in Figure~\ref{fig:limit}, the model incorrectly treats a hat as part of the background, resulting in visually unrealistic outputs. Additionally, due to complex background scenes, our model sometimes fails to produce temporally stable backgrounds, leading to visible flickering across views. These limitations stem from the fact that our current model focuses primarily on the facial region, without explicitly modeling or optimizing non-facial elements. We plan to explore these challenges more thoroughly in future work, expanding the generative scope of our model to better handle accessories, clothing, and backgrounds in a consistent and realistic manner.

\section{Social Impacts}
Our proposed NVB-Face model is designed to generate high-quality novel-view facial images from inputs of arbitrary quality. The original intent of this work is not for malicious use such as fraud or impersonation, but rather to address several challenging and socially beneficial tasks. For example, our model can be used to restore low-quality facial images and generate multi-view representations, thereby improving the performance of multi-view face recognition systems. In real-world scenarios, it could assist in transforming surveillance footage or low-resolution evidence frames into high-resolution multi-view facial images, enhancing the efficiency of suspect identification and identity verification. It also holds potential value for search-and-rescue efforts in locating missing persons through improved face matching. However, we are aware of potential misuse risks, such as applications in deepfake generation—e.g., creating high-quality multi-view facial avatars from a single social media image. To mitigate such risks, we propose to embed visible watermarks in the generated outputs to clearly indicate synthetic origin. We are committed to responsible AI development and will continue to keep the public informed of our latest updates, safeguards, and technical measures as our research evolves.

\end{document}